\ifdefined\XeTeXrevision\else
  \ifdefined\pdfoutput
    \pdfoutput=1
  \fi
\fi
\documentclass{article}
\usepackage{iclr2026_conference,times}
\iclrfinalcopy 

\usepackage{microtype}
\usepackage{graphicx}
\usepackage{booktabs}
\usepackage{amsmath}
\usepackage{amssymb} 
\usepackage{xcolor} 
\usepackage{caption}
\usepackage{enumitem}
\usepackage[colorlinks=true,linkcolor=black,citecolor=blue!50!black,urlcolor=blue!50!black]{hyperref}
\setcitestyle{numbers,square}

\definecolor{ordblue}{HTML}{2C6FAD}
\definecolor{signred}{HTML}{C4453C}

\newcommand{\rhoS}{\rho_{\mathrm{signed}}}
\newcommand{\rhoO}{\rho_{\mathrm{ordinal}}}

\title{Preference Is Not Intervention:\\
The Structure and Stability Boundaries of Reader-Specific Evidence Utility}

\author{Shi Zhou \\
College of Software, Jilin University \\
\texttt{zhoushi25@mails.jlu.edu.cn}}

\begin{document}
\maketitle
\lhead{Preprint} 

\begin{abstract}
\noindent ML systems increasingly condition decisions on downstream model
identity, but this is useful only if model-specific differences form reusable
structure rather than input-local interactions. We test this in
retrieval-augmented generation (RAG), where evidence utility can be measured
under controlled interventions. Holding query, evidence, task, scoring, and
intervention fixed, nine readers disagree on effect sign in 33\% of jointly
affected cells; reader$\times$query interaction explains 29.8\% of utility
variance versus an 8.4\% permutation null; and self-selected evidence improves
F1 by $+0.031$ ($t=3.39$). We then ask the sharper question:
\emph{which components of this heterogeneity are stable reader properties
across queries?} Separating three measurable objects---evidence
\emph{activity}, \emph{ordinal preference}, and \emph{conditional signed
direction}---we find ordinal reader geometry stable across four independent
settings (split-half $\rho=0.60$--$0.83$): leave-one-out interventions, PRISM
preferences, RAMDocs, and RAGuard. Signed geometry is task-bounded: weak in
open-ended QA (0.14, 0.35), especially for misleading and irrelevant evidence,
but strong in binary fact-checking (0.75) with no significant ordinal gap,
though still below its sparsity-matched ceiling. Sparsity, decoding noise, and
metric artifacts do not explain the main ordinal--signed gap. Finally, stable
ordinal similarity fails to predict cross-reader intervention transfer
(oracle-distance $\rho=-0.27$; regret reliability $-0.28$). Reader-specific
utility exists, but preference is not intervention: stable ranking similarity
does not license transfer of help/harm decisions.
\end{abstract}

\section{Introduction}

Modern ML systems increasingly condition their behavior on the identity of the
model that consumes their outputs: retrieval is personalized per generator,
queries are routed per model, ensembles are weighted per member. These bets
pay off only to the extent that observed model-specific differences contain
\emph{reusable structure}---a stable property of the model---rather than
situation-local interactions that evaporate on new inputs. We ask when this
holds, in a test bed where the question can be studied with unusual control:
evidence utility in retrieval-augmented generation (RAG).

Evidence selection in RAG is increasingly optimized
for \emph{downstream utility}: how much a document actually helps the
generator, rather than how relevant it looks to a retriever. This raises a
question the literature has not cleanly answered: does utility have a
systematic \textbf{reader-specific} component? It is established that
retriever preferences and LLM-friendly evidence diverge---the retriever--LLM
preference gap \citep{preferencegap2024}---and multi-system ranking work
personalizes retrieval for different RAG \emph{agents}
\citep{salemi2025multiagent}; but agents there differ simultaneously in task,
dataset, backbone, and strategy, so the effect of changing only reader identity is not
identified. What is missing is a controlled answer to: \emph{holding query,
evidence, task, and intervention fixed, does the identity of the reading model
itself induce reproducible differences in passage utility---and if so, what is
the structure of those differences?}

We answer both parts. \textbf{First, reader-specific utility is real.} On 9
readers and 100 NQ/HotpotQA queries with leave-one-out interventions, readers
disagree on whether a document helps or harms in 33\% of jointly-affected
cells (95\% CI [0.283, 0.377]); in 72\% of cases where a document moves at
least one reader of a pair, it moves exactly one; per-reader nonzero-utility
rates span 13--37\%; and a reader's own measured evidence selections beat the
average of other readers' selections by $+0.031$~F1 (query-clustered
$t = 3.39$). Reader identity is not a nuisance variable---it is a
load-bearing axis of evidence utility. A variance decomposition of the utility
tensor sharpens the picture: reader$\times$query interaction accounts for
29.8\% of utility variance versus a null median of 8.4\% ($p < 10^{-4}$),
whereas the reader main effect is only 0.4\%. Reader-involving terms sum to
68\% algebraically, but include an unreplicated three-way component (37.1\%)
that also absorbs residual variation. The dominant reader signal is thus
interaction with query--evidence context---the structure whose stability must
be established, not assumed.

\textbf{Second, reader specificity is not one thing.} Having established that
utilities differ across readers, we ask whether these differences constitute a
\emph{reusable reader property}---a question with direct implications for
persistent reader profiles and cross-reader transfer. We show that
``reader-specific utility'' decomposes into three measurable objects with
different cross-query stabilities: \emph{activity} (which documents move a
reader at all), \emph{ordinal preference} (how a reader ranks candidate
evidence), and \emph{conditional signed direction} (whether a document helps
or harms, among documents that move both readers). The decomposition matters
because different systems consume different objects: ranking-style supervision
measures the ordinal object, while inclusion decisions depend on the signed
one---a system could inherit ``reader preference'' structure without
inheriting anything about intervention direction.

That is exactly what we find. Ordinal reader geometry is \textbf{consistently
cross-query stable} in every setting we evaluate (split-half reliability
0.60/0.79/0.83/0.69 on our LOO arm, the independent PRISM preference data
\citep{rank4gen2026}, RAMDocs, RAGuard). Conditional signed geometry is
\textbf{substantially weaker and task-bounded}: near the floor in open-ended
QA (0.14 internal; 0.35 on RAMDocs)---far below a sparsity-matched
stable-world null---yet strongly stable in binary fact-checking (RAGuard 0.75,
no significant ordinal gap, though below its sparsity-matched ceiling): a boundary condition, localized
to misleading and irrelevant evidence (0.10/0.09) versus partially stable
supporting evidence (0.33).

\textbf{Third, the distinction has teeth downstream.} In a real cross-reader
transfer experiment (9$\times$9 source--target pairs, 50 pre-registered
queries, 4{,}050 cells), the reader specificity we establish does not assemble
into transferable structure: transfer regret is predicted neither by
behavioral profiles ($\rho = 0.05$) nor by in-sample \emph{oracle} utility
distance ($\rho = -0.27$, n.s.), and the regret matrix itself has no
split-half reliability ($-0.28$). Stable ordinal similarity---the thing that
\emph{is} stable---does not license cross-reader intervention decisions.

\paragraph{Contributions.}
\begin{itemize}[leftmargin=1.4em,itemsep=2pt]
\item \textbf{C1 --- Controlled characterization of reader-specific utility.}
Holding query, evidence, task, scoring, and intervention fixed, we show that
changing only reader identity yields substantial, structured differences in passage utility,
validated behaviorally by a significant self-reader selection advantage---to
our knowledge the first controlled, multi-reader characterization separating
the \emph{existence} of reader heterogeneity from the \emph{stability} of its
components (\S\ref{sec:heterogeneity}).
\item \textbf{C2 --- Stability decomposition and its boundary.} Evidence
activity and ordinal preference geometry are consistently cross-query stable
across four independent settings; conditional help/harm geometry is
substantially weaker, task-bounded, and localized to misleading/noise evidence
in open-ended QA; frozen calibrations show that sparsity, decoding, and metric
artifacts do not explain the main ordinal--signed gap
(\S\ref{sec:ordinal}--\S\ref{sec:localize}). A matched
forced-choice perturbation gives suggestive evidence that constraining the
answer space stabilizes responses to misleading evidence
(\S\ref{sec:mechanism}).
\item \textbf{C3 --- Practical consequence.} Stable ordinal reader similarity
does not provide a stable basis for cross-reader intervention transfer
(\S\ref{sec:transfer}).
\end{itemize}

\section{Related Work}
\label{sec:related}

\paragraph{Consumer-dependent retrieval and downstream utility.}
\citet{preferencegap2024} establish that retriever relevance preferences
diverge from what downstream LLMs can exploit; multi-agent ranking work
personalizes retrieval for 18 RAG agents from downstream feedback
\citep{salemi2025multiagent,salemi2024unified}; R3AG \citep{r3ag2026} routes
retrievers per generator. In all of these, consumers differ simultaneously in
task, dataset, backbone, and strategy, so reader identity is confounded with
task; we isolate it with everything else fixed.

\paragraph{LLM-specific utility notions.}
Concurrent preprint work proposes LLM-specific passage utility and reports
limited transfer of utilitarian passages across generators
\citep{llmspecific2025}. We independently establish reader-conditioned utility
differences under controlled interventions, but ask a different question: which
components of the heterogeneity constitute stable reader structure across
queries. Rank4Gen \citep{rank4gen2026} learns generator-conditioned ranking
from PRISM, which we use as an independent external replication resource for
the ordinal component. Generator-agnostic utility rerankers that transfer
across readers \citep{rrpo2026,lurerag2026} are consistent with our
decomposition: predictable utility lives largely on the query/evidence side and
in activity/ordinal structure.

\paragraph{Conflict robustness and methodology.}
RAMDocs \citep{ramdocs2025} and RAGuard \citep{raguard2025} supply typed
evidence (supporting / misleading / noise); MAGIC \citep{magic2025} benchmarks
inter-context conflict resolution. This literature asks whether models
\emph{answer correctly} under conflict; we ask whether the \emph{direction} of
a document's effect is a stable reader attribute---and find the instability
concentrates on exactly these adversarial evidence types. Our protocol builds
on classical split-half reliability with stratified splits and
sparsity-matched permutation calibration.

\section{Three Stabilities of Reader-Specific Utility}
\label{sec:framework}

\subsection{Readers and utility}

A \textbf{reader} is a model endpoint under a fixed deployment
configuration---model, decoding policy, and serving stack fixed for the
duration of the study---not an architecture-intrinsic personality. This makes
reader identity operational and is what transfer systems actually condition on
in deployment.

A \textbf{utility operator} maps (reader $m$, query $q$, document $d$) to a
scalar $U[m,q,d]$ relative to a baseline. We use two operators:
\begin{align}
\text{LOO:}\quad & U[m,q,d] = \mathrm{score}_m(q, D) - \mathrm{score}_m(q, D \setminus \{d\}),\\
\text{Single-doc:}\quad & U[m,q,d] = \mathrm{score}_m(q, \{d\}) - \mathrm{score}_m(q, \varnothing),
\end{align}
where $D$ is a top-$k$ retrieved context and $\varnothing$ is the closed-book
condition. Scores are deterministic task metrics (token-F1 or exact/binary
match, per dataset protocol).

\subsection{Three measurable objects}

\begin{figure}[t]
\centering
\includegraphics[width=0.58\linewidth]{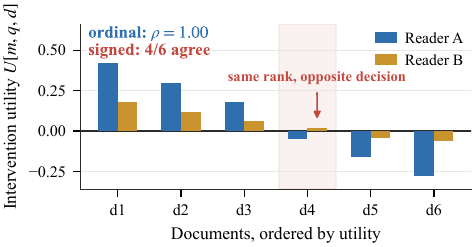}
\caption{\textbf{Same ranking, different zero-crossings (illustrative).}
Ordinal agreement between two readers can coexist with opposite help/harm
signs. Relative-preference supervision measures the former; inclusion
decisions depend on the latter.}
\label{fig:concept}
\end{figure}

From the utility tensor we derive three reader-pair geometries. For readers
$i, j$ and query $q$:
\begin{itemize}[leftmargin=1.4em,itemsep=1pt]
\item \textbf{Activity}: full-support sign agreement, counting the zero
pattern as signal---whether a document moves the reader at all.
\item \textbf{Ordinal preference}: Spearman correlation between the two
readers' utility vectors on $q$---relative preference, the object that
preference-pair and listwise supervision captures.
\item \textbf{Conditional signed direction}: sign agreement restricted to
documents nonzero for \emph{both} readers (dual-nonzero
cells)---intervention direction: help versus harm.
\end{itemize}
Each object induces a per-query distance between $i$ and $j$; aggregating over
queries (median) yields a reader-pair distance matrix $D$---the object's
\emph{reader geometry}. Figure~\ref{fig:concept} illustrates why the objects
must be separated: two readers can rank six documents identically (Spearman
$=1$) while disagreeing on the sign of a third of them.

\subsection{Stability as split-half reliability of the geometry}

An object's cross-query \textbf{stability} is the split-half reliability of
its geometry: split the queries into two stratified halves (by dataset source
and, where applicable, gold verdict), compute $D$ on each half, and correlate
the two distance vectors across reader pairs (Spearman $\rho$; 1{,}000 random
splits; we report the median and 2.5/97.5 percentiles). A stable reader
property yields $D_A \approx D_B$; a query-local interaction yields
uncorrelated halves. Pairwise geometry is the weakest object that
similarity-based personalization relies on, and it makes no parametric
commitment about reader profiles; whether it suffices for profile-level
prediction is tested directly in \S\ref{sec:transfer}.

\subsection{Permutation calibration against sparsity}
\label{sec:calibration}

Signed utility is sparse: most documents do not change most readers' scores,
so dual-nonzero cells are rare, and \emph{some} loss of split-half reliability
is expected from sparsity alone. To separate measurement sparsity from genuine
instability, every signed estimate is calibrated against an arm-specific
\textbf{stable-world permutation}. All variants preserve the observed sparse
support and relevant conflict-count marginals while imposing stable
reader-pair conflict propensities across queries. The internal arm permutes
observed conflict indicators within each reader pair; the external arms
permute them within reader-pair$\times$evidence-position strata (2{,}000--
5{,}000 simulations). Appendix~\ref{app:nulls} gives both constructions.
Observed $\rho$ far below the null median rejects ``sparsity explains the
weakness.'' A parallel null calibrates ordinal stability against type-level
structure alone; repeated-decoding runs bound decoding stochasticity
(\S\ref{sec:robust}).

\subsection{What stability would license}

If conditional signed geometry were stable, a reader's help/harm pattern
estimated on one query set would transfer to new queries, reader profiles
could be built once and reused, and reader-conditioned selection could target
intervention direction. If only ordinal geometry is stable, ranking-style
reader conditioning may still work---but inclusion decisions and cross-reader
transfer of intervention choices are not licensed by the same evidence.
Section~\ref{sec:transfer} tests the transfer consequence directly.

\section{Experimental Setup}
\label{sec:setup}

\subsection{Readers}

The internal LOO arm uses \textbf{9 core readers}: five API endpoints
(Qwen3.6-Flash, DeepSeek-V4-Flash, GLM-5.2, GPT-5.6-Luna, K3) and four local
8--9B GGUF endpoints (Qwen3.5-9B-Instruct, Ministral-8B-Instruct,
Llama-3.3-8B-Instruct, Llama-3.1-8B-Instruct) served on a single RTX-4060
host (full roster and decoding configurations: Appendix~\ref{app:roster}). The external single-document arm uses \textbf{13 readers}: the 9 core
plus Qwen3.7-Plus, Qwen3.7-Max, Qwen3.8-Max, and DeepSeek-V4-Pro. Decoding is
deterministic wherever the endpoint permits (one endpoint requires temperature
1.0 with forced reasoning; it is treated as a distinct deployment
configuration per \S\ref{sec:framework}). Local endpoints are quantized
builds; the design compares \emph{relative geometry across readers}, not
absolute capability tiers, and scale-strata cuts are reported as descriptive
only.

\subsection{Internal arm: LOO utility on NQ/HotpotQA}

100 queries (50 NQ \citep{nq2019}, 50 HotpotQA \citep{hotpotqa2018}), each
with 8 unique BGE-M3-ranked \citep{bgem3} candidate documents with guaranteed
supporting evidence. Every reader answers 10 conditions per query
(closed-book, full 8-document context, and eight leave-one-out contexts),
scored by token-level F1 against gold answers: a $9 \times 100 \times 8$
utility tensor (7{,}200 cells). Prompt templates and scoring contracts for
all arms are given in Appendix~\ref{app:prompts}.

\subsection{External arms: single-document interventions}

\textbf{RAMDocs} \citep{ramdocs2025} (open-ended disambiguation QA with
misinformation): 149 queries meeting a frozen eligibility rule---at least two
supporting (correct), one misleading (misinformation), and one noise
document---with documents selected by frozen data order. Each (reader, query)
contributes 5 conditions: closed-book plus four single-document contexts
(support1, support2, mislead1, noise1). Scoring follows the dataset's strict
protocol, adapted to single-document conditions as normalized any-gold match.

\textbf{RAGuard} \citep{raguard2025} (real-world fact verification against
misleading Reddit retrievals): 212 claims under the same eligibility rule;
scoring is binary verdict match with deterministic parsing. Both arms stratify
query splits by dataset and (for RAGuard) gold verdict. Per-position signed
analyses pool the two support positions and report the misleading and noise
positions separately.

\subsection{External ordinal arm: PRISM}

PRISM/Rank4Gen-DPO public data \citep{rank4gen2026}: 58{,}404 preference rows
$\to$ 29{,}197 unique (query, generator) keys (bilingual prompt variants
deduplicated), covering 7{,}791 unique queries across five sources (HotpotQA,
2WikiMultiHopQA, MuSiQue, MS~MARCO, CRUD-RAG) and seven downstream generators.
Each key's \emph{chosen} ordered document set is the generator's preferred
context. Candidate indices are translated to per-query canonical document ids
via normalized-text digests (99.98\% pool-Jaccard across generators), so
cross-generator comparisons are of the same documents. Ordinal geometry:
per-query RBO \citep{rbo2010} ($p = 0.9$) between chosen ordered sequences;
composition-only auxiliary via Jaccard; 21 generator pairs; the same
1{,}000-split stratified protocol. PRISM has no signed operator---it
calibrates the ordinal half only.

\paragraph{Freezing and reproducibility.}
Query eligibility rules, per-arm decision rules, split counts and seeds,
parsing tiers, and the transfer plan were frozen before results analysis;
artifact-level checksums and frozen plans are retained
(\S\ref{sec:artifacts}). All analyses run on frozen artifacts with fixed
seeds.

\section{Results}
\label{sec:results}

\subsection{RQ1 --- Reader-specific utility is real (controlled heterogeneity)}
\label{sec:heterogeneity}

\begin{figure}[t]
\centering
\includegraphics[width=0.96\linewidth]{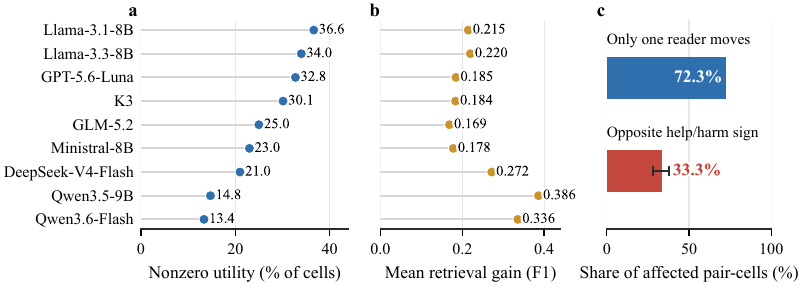}
\caption{\textbf{Changing only the reader changes which evidence matters.}
Across 9 readers and 100 fixed NQ/HotpotQA queries, (a) activity spans
13.4--36.6\%, (b) mean retrieval gain is positive for every reader, and (c)
72.3\% of affected pair-cells move only one reader while 33.3\% reverse the
help/harm sign when both move.}
\label{fig:heterogeneity}
\end{figure}

Before stability, existence. With everything else fixed, changing only reader identity
yields substantial differences (Figure~\ref{fig:heterogeneity}; per-reader
statistics in Appendix~\ref{app:perreader}). All readers
benefit from retrieval on average (gain $+0.169$ to $+0.386$ F1), but
\emph{which} documents drive the gain differs sharply:

\begin{itemize}[leftmargin=1.4em,itemsep=2pt]
\item \textbf{Sign disagreement.} Pooling all 36 reader pairs, readers assign
\emph{opposite signs} to a document's effect in \textbf{33.3\%} of
dual-nonzero cells (1{,}067/3{,}206; query-cluster bootstrap 95\% CI
[0.283, 0.377]).
\item \textbf{Activity asymmetry.} In \textbf{72.3\%} of (pair, cell) cases
where a document moves at least one reader of a pair, it moves exactly one
(8{,}348/11{,}554). Per-reader nonzero rates span 13.4--36.6\%
(2.7$\times$). The external arms show the same spread (RAGuard:
16.8--51.5\%).
\item \textbf{Behavioral validation.} A reader's own measured preference set
outperforms the average of other readers' sets by \textbf{$+0.031$~F1}
(query-clustered $t = 3.39$, df $= 49$; in sample, \S\ref{sec:transfer}) ---
reader-specific utility is not only measurable but actionable in principle.
\end{itemize}

A three-way variance decomposition of the 7{,}200-cell tensor (reader $\times$
query $\times$ rank position; Appendix~\ref{app:perreader}) makes the same
point at the distribution level: the clean reader$\times$query interaction
accounts for 29.8\% of variance versus a permutation-null median of 8.4\%
($p < 10^{-4}$), while the reader main effect is only 0.4\%. All
reader-involving terms sum to 68.0\% of the algebraic decomposition. Because
there is one observation per reader$\times$query$\times$position cell, that
sum includes a 37.1\% three-way component that also absorbs residual
variation. The signal is nevertheless large and interaction-led, not a
uniform reader-level shift.

Heterogeneity of this size is what reader-conditioned evidence selection hopes
to exploit. The rest of the paper asks what part of it is a stable reader
property.

\subsection{RQ2 --- Ordinal preference geometry is consistently stable}
\label{sec:ordinal}

\begin{figure}[t]
\centering
\includegraphics[width=0.88\linewidth]{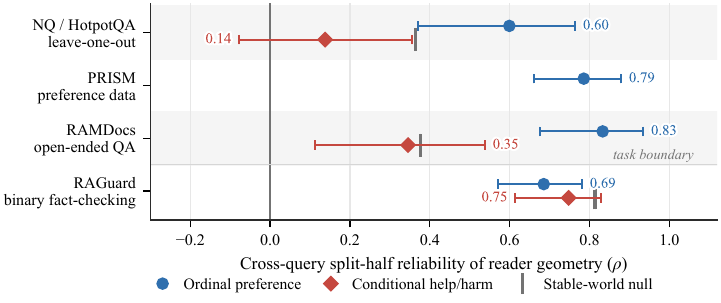}
\caption{\textbf{Ordinal stability is broad; signed stability has a task
boundary.} Cross-query split-half reliability of reader-pair geometry across
four independent settings (points: medians over 1{,}000 stratified splits;
whiskers: 2.5/97.5 percentiles). Blue: ordinal geometry. Red: conditional
signed geometry. Gray ticks: sparsity-matched stable-world nulls
(\S\ref{sec:calibration}). PRISM has no signed operator.}
\label{fig:forest}
\end{figure}

Ordinal reader geometry reproduces across disjoint query halves in every
evaluated setting (Figure~\ref{fig:forest}, blue): 0.599 [0.370, 0.764] on the
internal LOO arm, 0.786 [0.660, 0.879] on PRISM, 0.833 [0.676, 0.934] on
RAMDocs, and 0.685 [0.571, 0.781] on RAGuard---spanning two utility operators,
two answer formats, and an independent preference pipeline with different
generators and query sources. On PRISM the result survives artifact controls:
an identity-shuffle null (chosen-list lengths preserved, document identities
resampled) yields only 0.314 against the observed 0.786; an order-shuffle null
yields 0.703, so composition carries most of the stability and ordering adds
$\sim$0.08; an exactly size-matched comparison is \emph{more} stable, not less
(0.862); and composition-only Jaccard geometry is likewise stable (0.742).
Full split distributions and null calibrations are in
Appendices~\ref{app:splits} and~\ref{app:nulls}.

\subsection{RQ3 --- Conditional signed geometry is weaker, with a task boundary}
\label{sec:boundary}

\begin{table}[t]
\centering\small
\begin{tabular}{lcccc}
\toprule
Setting & $\rhoO$ & $\rhoS$ [95\% range] & Stable null & $p$ \\
\midrule
NQ/HotpotQA LOO (9 readers) & 0.599 & \textbf{0.138} [$-0.077$, 0.356] & 0.363 & $2{\times}10^{-4}$ \\
RAMDocs single-doc (13 readers) & 0.833 & \textbf{0.345} [0.113, 0.538] & 0.376 & $5{\times}10^{-4}$ \\
RAGuard single-doc (13 readers) & 0.685 & \textbf{0.748} [0.614, 0.829] & 0.814 & $5{\times}10^{-4}$ \\
\bottomrule
\end{tabular}
\caption{\textbf{Conditional signed vs.\ ordinal stability and stable-world
nulls.} In open-ended QA, signed geometry lies far below ordinal geometry and
its sparsity-matched ceiling. In binary fact-checking it is statistically
indistinguishable from ordinal geometry, although still below that ceiling.}\label{tab:signed}
\end{table}

Three observations (Figure~\ref{fig:forest}, red; Table~\ref{tab:signed}).
\textbf{First}, in open-ended QA the conditional signed geometry is far below
the ordinal geometry---paired per-split $\Delta = 0.487$ [0.226, 0.721] on
RAMDocs---and far below what measurement sparsity alone predicts: the
stable-world null expects $\rho \approx 0.36$--$0.38$ under identical support,
while observed values are 0.14 and 0.35 ($p = 2\!\times\!10^{-4}$ and
$5\!\times\!10^{-4}$; two further calibrations concur,
Appendix~\ref{app:nulls}). \textbf{Second}, the weakness is not universal: in
binary fact-checking the signed geometry is as stable as the ordinal one
(paired $\Delta = -0.064$ [$-0.201$, 0.114]). It is strongly stable, not at
the stable-world ceiling (0.748 vs.\ 0.814, $p=5\!\times\!10^{-4}$). Our own
data therefore reject the universal instability claim and instead establish a \textbf{task
boundary}. \textbf{Third}, both external arms use a different intervention
operator than the internal arm, so the open-QA weakness is not an artifact of
leave-one-out redundancy. The signed reading is ``substantially weaker than
ordinal and below its sparsity-matched ceiling,'' not ``zero.''

\subsection{RQ4 --- In open-ended QA, instability localizes to misleading and noise evidence}
\label{sec:localize}

\begin{figure}[t]
\centering
\includegraphics[width=0.82\linewidth]{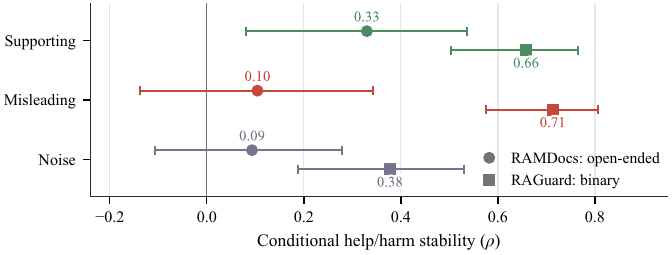}
\caption{\textbf{The task boundary localizes to evidence type.} Points and
whiskers show per-position signed split-half reliability and its 2.5/97.5
percentiles. RAMDocs (circles) is weak for misleading and noise evidence;
RAGuard (squares) is stable across all three positions.}
\label{fig:perposition}
\end{figure}

RAMDocs per-position signed stability (Figure~\ref{fig:perposition}):
supporting evidence 0.330 [0.080, 0.536] over 14{,}549 dual-nonzero cells;
misleading 0.104 [$-0.137$, 0.343] over 2{,}561; noise 0.093 [$-0.107$, 0.279]
over 1{,}888. The weakness is not uniform across evidence processing:
directions of \emph{supporting} evidence are partially stable, while
directions of \emph{misleading} and \emph{irrelevant} evidence are essentially
query-local. RAGuard shows the opposite pattern (0.658 / 0.713 / 0.377)---in
binary verification, even misleading-evidence direction is stable across
queries. What varies by task regime is specifically the cross-query stability
of \textbf{how a reader responds to adversarial evidence}.

\subsection{A matched mechanism probe --- forced choice (suggestive)}
\label{sec:mechanism}

\begin{figure}[t]
\centering
\includegraphics[width=0.88\linewidth]{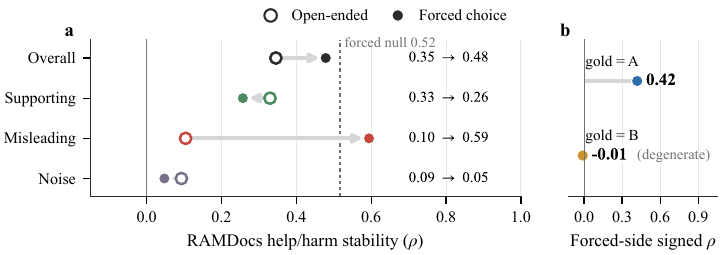}
\caption{\textbf{Constraining the answer space (matched perturbation, RAMDocs
arm).} (a)~Open-ended vs.\ forced-choice signed stability, overall and by
evidence type: the elevation concentrates on misleading evidence ($0.10 \to
0.59$). (b)~Label-stratum asymmetry: stabilizing geometry measurable in the
gold\,=\,A stratum; gold\,=\,B degenerate.}
\label{fig:forced}
\end{figure}

One candidate mechanism is the answer space: binary verdicts leave less room
for reader-specific conflict resolution than open-ended answers. We test this
with a matched RAMDocs perturbation that changes only the output contract to a
two-option forced choice (gold vs.\ misinformation-supported answer; fixed
A/B assignment), using the pre-registered decision rule in Appendix~\ref{app:labels}.

Overall signed stability rises from 0.345 to 0.479 (paired $\Delta=+0.130$
[$-0.114$, 0.403]), partial evidence under the frozen rule. The elevation is
misleading-selective (0.104 to 0.594), while supporting evidence is unchanged
(0.330 to 0.257). The forced-side mislead geometry is measurable only in the
gold\,=\,A stratum (0.599; gold\,=\,B degenerate), and remains below its own
stable null (0.516, $p=5\!\times\!10^{-4}$). We therefore treat the answer-space
account as a \textbf{suggestive mechanism}, not an identified cause; the task
boundary itself is the established result.

\subsection{Practical consequence --- stable similarity does not license transfer}
\label{sec:transfer}

\begin{figure}[t]
\centering
\includegraphics[width=0.88\linewidth]{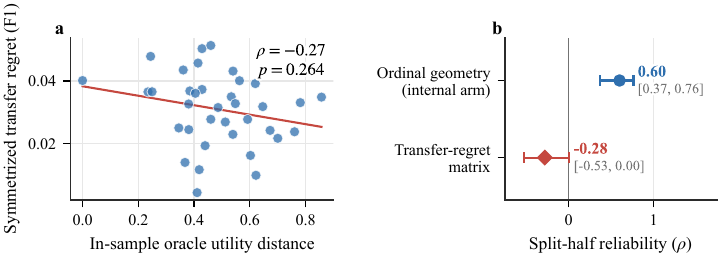}
\caption{\textbf{Stable similarity does not predict intervention transfer.}
(a) Across 36 reader pairs, in-sample oracle utility distance is unrelated to
symmetrized transfer regret ($\rho=-0.27$, $p=0.264$). (b) The regret matrix
has no reliable split-half structure ($\rho=-0.28$, 95\% interval
[$-0.53$, $0.00$]); self-selected evidence still gives the in-sample
$+0.031$ F1 advantage reported in the text.}
\label{fig:transfer}
\end{figure}

If stable ordinal geometry licensed intervention-level personalization, reader
similarity should predict transfer. We test this on 9 target readers and 50
pre-registered queries using tie-aware source preference sets, yielding 4{,}050
source$\times$target$\times$query cells (construction details:
Appendix~\ref{app:transfer}).

\textbf{Reader-specific utility is real (in sample):} self-selected evidence
beats other readers' sets by $+0.031$~F1 ($t=3.39$). \textbf{But similarity
does not predict transfer:} oracle utility distance vs.\ regret gives
$\rho=-0.271$ ($p=0.264$), and the regret matrix has no split-half reliability
($-0.281$ [$-0.530$, 0.001]; mean regret 0.035~F1, mean standard error 0.037).

Stable ordinal similarity is therefore not a sufficient basis for cross-reader
intervention decisions, even when similarity is measured in sample.

\section{Alternative Explanations}
\label{sec:robust}

\begin{table}[t]
\centering\footnotesize
\setlength{\tabcolsep}{3pt}
\begin{tabular}{p{3.1cm}p{10.4cm}}
\toprule
Alternative & Evidence and disposition \\
\midrule
Utility sparsity inflates instability & Stable-world permutation holding support fixed: \textbf{rejected}; $\rhoS$ remains far below the null (LOO 0.138 vs.\ 0.363; RAMDocs 0.345 vs.\ 0.376, both $p{\le}5{\times}10^{-4}$). \\
Decoding stochasticity & Repeated-decoding test--retest and attenuation bound: \textbf{rejected as sole cause}; $f{=}0.123$ caps corrected $\rhoS$ at 0.242, far below $\rhoO=0.599$. \\
Metric/threshold artifact & Matched metrics on identical support: localized; full-support activity is stable (0.650 $\approx$ 0.599), while signed direction is weak (0.138). \\
PRISM measures output habits & Identity-shuffle, order-shuffle, and size-matched nulls: \textbf{rejected}; identity null 0.314 vs.\ 0.786, order null 0.703, and size-matched control \emph{raises} stability to 0.862. \\
Leave-one-out operator artifact & Replication under a single-document operator: open-QA weakness persists across operators (RAMDocs). \\
Universal signed instability & External binary fact-checking check: \emph{rejected by our own data} (RAGuard 0.748); the task boundary remains. \\
Answer-space mechanism & Matched forced-choice perturbation: partial evidence ($\Delta=+0.130$, CI crosses 0); mislead-selective elevation is measurable in the gold\,=\,A stratum only. \\
\bottomrule
\end{tabular}
\caption{\textbf{Alternative explanations and their disposition.}}
\label{tab:robust}
\end{table}

Table~\ref{tab:robust} summarizes the controls. The test--retest analysis
(Appendix~\ref{app:retest}) shows that decoding stochasticity is real but does
not explain the main ordinal--signed gap: the attenuation-corrected ceiling
remains far below ordinal stability on both the internal and RAMDocs arms.

\section{Discussion}
\label{sec:discussion}

\paragraph{What kind of reader specificity exists?}
Our results separate three claims that ``reader-specific utility'' collapses
into one. Readers genuinely differ on matched query--evidence conditions, and consequentially
(\S\ref{sec:heterogeneity}). The \emph{ordinal} reading is additionally
stable: it reproduces across disjoint query sets in all four settings,
including an independent preference pipeline. The \emph{signed} reading is
not uniformly stable: weak in open-ended QA, concentrated on adversarial
evidence, strong in binary verification. ``Utility is reader-specific'' is
true per query, yet not a stable signed trait across queries in open-ended QA.

\paragraph{Reconciling generator-conditioned and generator-agnostic methods.}
Generator-conditioned ranking can work because ordinal/compositional
preference geometry is genuinely cross-query stable; generator-agnostic
utility rerankers can generalize because much predictable utility lives on
the query/evidence side and in activity structure. Both are compatible with
signed direction being query-local: neither ranking nor transferable utility
estimation requires stable help/harm geometry.

\paragraph{For personalization systems.}
Condition on the reader for \emph{ranking and composition}, where stability
holds. For \emph{intervention-level} personalization, stable ordinal
similarity implies neither transferable intervention decisions nor a reliably
structured transfer outcome: treat help/harm direction as query-local unless
the task regime is known to stabilize it.

\paragraph{For evaluation practice.}
Preference-pair and listwise supervision measure ordinal structure; deployment
decisions depend on intervention direction. A method can look stable under
ranking-style evaluation while the property that matters for inclusion
decisions is query-local. Cross-query reliability of \emph{signed}
utility---with sparsity-matched calibration---should accompany any claim of
stable reader preference. Our data suggest the boundary hypothesis that
intervention direction stabilizes when the task constrains how evidence can be
realized in the answer; identifying the causal axis is left to future work.

\section{Limitations}
\label{sec:limitations}

The task boundary is established, but its causal axis is not identified: the
forced-choice perturbation is partial evidence with label-stratum degeneracy.
We do not claim priority for model-dependent passage utility (a concurrent
preprint proposes the notion \citep{llmspecific2025}); ours is the controlled
characterization and stability decomposition. The two arms use different
reader panels and PRISM has no signed operator; cross-setting comparisons are
of stability patterns, not identical measurements. Transfer uses 50 queries
per cell; utility is task-metric-based.

\section{Conclusion}

Reader-specific evidence utility is real and consequential: with all else
fixed, changing the reader changes which documents help, harm, or do nothing.
Relative evidence preferences are consistently
cross-query stable across readers, datasets, and preference constructions; the
stability of help-versus-harm directions is task-bounded---weaker in
open-ended QA, concentrated on misleading and irrelevant evidence, and absent
as a gap in binary fact-checking. Preference is not intervention: stable
ordinal geometry licenses neither cross-reader transfer nor claims of stable
help/harm direction.

{\small
\bibliography{references}
\bibliographystyle{plainnat}
}

\section*{Ethics and Reproducibility}

All datasets are public (NQ, HotpotQA, RAMDocs, RAGuard, PRISM) and used
within their licenses; no human subjects or personal data are involved.
RAGuard documents originate from Reddit; we use them only as retrieved
evidence under the dataset's terms. All experimental decisions reported here
were frozen before results analysis; artifacts, frozen plans, and analysis
scripts will be released. Misleading-evidence findings describe model behavior
under misinformation and carry no endorsement of the misleading content.

\textbf{Provenance.} An earlier candidate reader failed the response-validity
criterion and was excluded before the final analysis panel; primary
conclusions were recomputed on the finalized roster.

\section*{AI Usage Disclosure}

AI assistants were used for literature cross-checking, feedback on framing and
experimental design, interpretation of results, and manuscript editing. All
experimental decision rules were frozen before results analysis, all analyses
were executed programmatically on frozen artifacts, and every reported number
was recomputed from those artifacts. The authors take full responsibility for
all content. 

\appendix

\section{Reader roster and deployment configurations}
\label{app:roster}

Table~\ref{tab:roster} lists the 13 readers. Following
\S\ref{sec:framework}, a \emph{reader} is a model endpoint under a fixed
deployment configuration. All endpoints except K3 run deterministic decoding
(temperature 0, reasoning disabled, 128-token cap). K3's endpoint forces
temperature 1.0 and always-on reasoning; we run it with the minimal reasoning
effort and a 2{,}048-token budget and analyze it as its own deployment
configuration. Local readers are GGUF builds served on a single RTX-4060
host. Internal identifiers predate the final display names:
\texttt{reader\_qwen\_flash}, \texttt{reader\_deepseek\_flash}, and
\texttt{reader\_glm} denote Qwen3.6-Flash, DeepSeek-V4-Flash, and GLM-5.2
respectively.

\begin{table}[h]
\centering\footnotesize
\setlength{\tabcolsep}{4pt}
\resizebox{\linewidth}{!}{%
\begin{tabular}{llllc}
\toprule
Reader & Endpoint model & Deployment & Decoding & Arms \\
\midrule
Qwen3.6-Flash & \texttt{qwen3.6-flash} & hosted API & T${=}0$, no reasoning, 128 tok & LOO, SD \\
DeepSeek-V4-Flash & \texttt{deepseek-v4-flash-0731} & hosted API & T${=}0$, no reasoning, 128 tok & LOO, SD \\
GLM-5.2 & \texttt{glm-5.2} & hosted API & T${=}0$, no reasoning, 128 tok & LOO, SD \\
GPT-5.6-Luna & \texttt{gpt-5.6-luna} & hosted API & T${=}0$, reasoning off, 128 tok & LOO, SD \\
K3 & \texttt{k3-256k} & hosted API & T${=}1.0^{*}$, reasoning low$^{*}$, 2048 tok & LOO, SD \\
Qwen3.5-9B-Instruct & Q3\_K\_M GGUF & local & T${=}0$, no reasoning, 128 tok & LOO, SD \\
Ministral-8B-Instruct-2410 & Q4\_K\_M GGUF & local & T${=}0$, no reasoning, 128 tok & LOO, SD \\
Llama-3.3-8B-Instruct & Q4\_K\_M GGUF & local & T${=}0$, no reasoning, 128 tok & LOO, SD \\
Llama-3.1-8B-Instruct & Q4\_K\_M GGUF & local & T${=}0$, no reasoning, 128 tok & LOO, SD \\
Qwen3.7-Plus & \texttt{qwen3.7-plus} & hosted API & T${=}0$, no reasoning, 128 tok & SD \\
Qwen3.7-Max & \texttt{qwen3.7-max} & hosted API & T${=}0$, no reasoning, 128 tok & SD \\
Qwen3.8-Max & \texttt{qwen3.8-max} & hosted API & T${=}0$, no reasoning, 128 tok & SD \\
DeepSeek-V4-Pro & \texttt{deepseek-v4-pro} & hosted API & T${=}0$, no reasoning, 128 tok & SD \\
\bottomrule
\end{tabular}
}
\caption{\textbf{Reader roster.} Arms: LOO $=$ internal leave-one-out arm (9
readers); SD $=$ external single-document arms (13 readers).
$^{*}$Endpoint constraint: K3 permits neither lower temperatures nor
disabling reasoning; it is analyzed as a distinct deployment configuration.}
\label{tab:roster}
\end{table}

The pre-registered exclusion rule (fallback/parse failure $>25\%$ of a
reader's conditions, or nonzero utility rate $<5\%$) was applied mechanically
per arm. No reader triggered it in any final arm; the highest fallback rate
is 21.6\% (Ministral-8B on RAMDocs) and the next highest 7.8\%. Per-reader QC
statistics are recorded in the analysis artifacts
(Appendix~\ref{sec:artifacts}). Scale strata used for descriptive cuts on the
single-document arm: \emph{api-mid} $=$ \{Qwen3.6-Flash, Qwen3.7-Plus,
DeepSeek-V4-Flash\}; \emph{frontier} $=$ \{Qwen3.7-Max, Qwen3.8-Max,
DeepSeek-V4-Pro, GLM-5.2, GPT-5.6-Luna, K3\}; \emph{local-small} $=$ the four
local readers.

\section{Prompts and scoring contracts}
\label{app:prompts}

All QA-style conditions---every LOO condition, the RAMDocs and RAGuard
single-document conditions, and all transfer evaluations---share one prompt
contract:

\begin{quote}
\footnotesize
\textbf{System:} You are answering a factual question. Follow the output
format exactly.\\[4pt]
\textbf{User:} Question: $\langle$\textit{question}$\rangle$\\
Evidence: [1] $\langle$\textit{document 1}$\rangle$; [2]
$\langle$\textit{document 2}$\rangle$; \ldots; [$k$]
$\langle$\textit{document k}$\rangle$\\
Return only: Answer: $\langle$\textit{short answer}$\rangle$
\end{quote}

Closed-book conditions replace the evidence block with ``\texttt{(No evidence
provided.)}''. RAGuard questions are wrapped as ``Is the following claim true
or false? Claim: `$\langle$\textit{claim}$\rangle$'\,'' with gold answers
\texttt{True}/\texttt{False}; document text is the title plus the first
4{,}000 characters of the full text (97.2\% of selected documents covered in
full).

\textbf{Forced-choice menu} (\S\ref{sec:mechanism}; RAMDocs arm only):

\noindent\begin{minipage}{\linewidth}
\begin{quote}
\footnotesize
$\langle$\textit{question}$\rangle$\\
Which of the two options is the correct answer to the question above?\\
A) $\langle$\textit{option A}$\rangle$\\
B) $\langle$\textit{option B}$\rangle$\\
Respond with the option letter only.
\end{quote}
\end{minipage}

Option A is the supporting answer and option B the misinformation answer iff
the md5 hash of the query id is even---deterministic and computable from the
query id alone. In 44 of 149 rows the two supporting documents assert
different valid interpretations; the menu gold then follows support1, frozen
uniformly across readers.

\textbf{Scoring.} Answers are normalized (lowercase; punctuation stripped;
articles removed; whitespace collapsed) and scored as the maximum over gold
aliases, as exact match or token-level F1. The prediction is the last line
matching \texttt{answer:}; failing that, the last nonempty line (a leading
``final answer:'' is stripped). Forced-choice responses use a tiered parser:
(1)~exact single letter; (2)~leading letter with separator and matching
option text; (3)~unambiguous option-text mention; otherwise tier-0, scored
incorrect (per-reader tier-0 rates 0.0--2.6\%).

\textbf{Utility operators.} LOO arm: $U(m,q,d) = \mathrm{F1}(\text{full
8-document context}) - \mathrm{F1}(\text{context minus } d)$. Single-document
arms: $U(m,q,d) = \mathrm{score}(\text{document } d \text{ only}) -
\mathrm{score}(\text{closed-book})$, with score $=$ F1 (open-ended) or
accuracy (forced choice).

\section{Per-reader utility statistics (LOO arm)}
\label{app:perreader}

\begin{table}[h]
\centering\footnotesize
\begin{tabular}{lcc}
\toprule
Reader & Nonzero utility rate & Mean full-context gain (F1) \\
\midrule
Qwen3.6-Flash & 0.134 & $+0.336$ \\
DeepSeek-V4-Flash & 0.210 & $+0.272$ \\
GLM-5.2 & 0.250 & $+0.169$ \\
GPT-5.6-Luna & 0.328 & $+0.185$ \\
K3 & 0.301 & $+0.184$ \\
Qwen3.5-9B-Instruct (Q3) & 0.148 & $+0.386$ \\
Ministral-8B-Instruct (Q4) & 0.230 & $+0.178$ \\
Llama-3.3-8B-Instruct (Q4) & 0.340 & $+0.220$ \\
Llama-3.1-8B-Instruct (Q4) & 0.366 & $+0.215$ \\
\bottomrule
\end{tabular}
\caption{\textbf{Per-reader utility statistics on the internal LOO arm} (800
cells per reader: 100 queries $\times$ 8 documents). Nonzero rate: fraction
of cells with $U \neq 0$. Gain: $\mathrm{F1}(\text{full context}) -
\mathrm{F1}(\text{closed-book})$, averaged over queries.}
\label{tab:perreader}
\end{table}

Pooling all 36 reader pairs: dual-nonzero sign conflicts occur in 1{,}067 of
3{,}206 cells (33.3\%; query-cluster bootstrap 95\% CI [0.283, 0.377];
10{,}000 resamples over the 100 queries); activity asymmetry in 8{,}348 of
11{,}554 pair--cell cases (72.3\%). The dual-nonzero support of 3{,}206 is
pinned as the checksum of the sparsity calibration
(Appendix~\ref{app:nulls}).

\textbf{Variance decomposition.} Treating the tensor as a balanced
reader $\times$ query $\times$ position layout (one observation per cell), the
exact sums-of-squares decomposition attributes variance as follows
(query-cluster bootstrap 95\% CI, 10{,}000 resamples): reader main effect
0.4\% [0.3, 1.4]; query 6.0\% [4.2, 7.9]; position 2.3\% [1.1, 4.5];
reader$\times$query 29.8\% [25.7, 33.0]; reader$\times$position 0.8\% [0.8,
1.6]; query$\times$position 23.7\% [19.8, 27.6]; reader$\times$query$\times$
position (residual) 37.1\% [33.0, 40.6]. All terms involving the reader sum to
68.0\% [63.4, 72.5] in this algebraic decomposition. Because every
reader$\times$query$\times$position cell is observed once, the three-way term
cannot be separated from measurement error or other residual variation; the
68.0\% sum must be read with that qualification. Permutation tests (10{,}000 simulations; values shuffled
across readers within each query--position column, destroying all reader
structure) reject the null for the reader main effect (observed 0.4\% vs.\
null median 0.08\%, $p < 10^{-4}$) and for reader$\times$query (29.8\% vs.\
8.4\%, $p < 10^{-4}$), but not for reader$\times$position (0.8\% vs.\ 0.6\%,
$p = 0.08$). The clean result is therefore the large reader$\times$query
interaction, not a uniform reader shift; the 68.0\% reader-involving sum is a
broader descriptive total.

\section{Full split distributions}
\label{app:splits}

All estimates use the frozen stratified split-half protocol (1{,}000 splits;
seeds 20260815 internal/PRISM, 20260817 external arms).
Table~\ref{tab:splits} reports the full split distributions underlying
Figure~\ref{fig:forest} and Table~\ref{tab:signed};
Table~\ref{tab:perposition} the per-position breakdown underlying
Figures~\ref{fig:perposition} and~\ref{fig:forced}. On the internal arm,
80.6\% of ordinal splits exceed 0.5 while 92.5\% of signed splits fall below
0.3 (11.5\% negative).

\begin{table}[tb]
\centering\footnotesize
\setlength{\tabcolsep}{4pt}
\begin{tabular}{llcccccc}
\toprule
Setting & Object & 2.5 & 25 & Median & 75 & 97.5 & NN \\
\midrule
NQ/HotpotQA LOO & ordinal & 0.370 & 0.525 & 0.599 & 0.664 & 0.764 & 0.397 \\
 & signed & $-$0.077 & 0.058 & 0.138 & 0.218 & 0.356 & 0.198 \\
PRISM (RBO) & ordinal & 0.660 & 0.748 & 0.786 & 0.819 & 0.879 & 0.571 \\
PRISM (Jaccard) & ordinal & 0.445 & 0.680 & 0.742 & 0.836 & 0.999 & 0.991 \\
RAMDocs & ordinal & 0.676 & 0.779 & 0.833 & 0.870 & 0.934 & 0.884 \\
 & signed & 0.113 & 0.268 & 0.345 & 0.417 & 0.538 & 0.246 \\
 & paired $\Delta$ & 0.226 & 0.403 & 0.487 & 0.566 & 0.721 & --- \\
RAGuard & ordinal & 0.571 & 0.646 & 0.685 & 0.721 & 0.781 & 0.361 \\
 & signed & 0.614 & 0.708 & 0.748 & 0.784 & 0.829 & 0.402 \\
 & paired $\Delta$ & $-$0.201 & $-$0.108 & $-$0.064 & $-$0.012 & 0.114 & --- \\
RAMDocs forced & signed & 0.290 & 0.414 & 0.479 & 0.536 & 0.624 & --- \\
 & $\Delta_{\mathrm{format}}$ & $-$0.114 & 0.039 & 0.130 & 0.214 & 0.403 & --- \\
 & ordinal (descr.) & 0.437 & 0.674 & 0.748 & 0.808 & 0.900 & --- \\
\bottomrule
\end{tabular}
\caption{\textbf{Split-half reliability distributions} (percentiles over
1{,}000 splits; 999 for forced ordinal). NN: mean fraction of readers whose
nearest neighbor in one half's distance matrix is also nearest in the other
half's. Paired $\Delta$ rows are per-split ordinal $-$ signed differences;
$\Delta_{\mathrm{format}}$ is the per-split forced $-$ open signed
difference.}
\label{tab:splits}
\end{table}

\begin{table}[tb]
\centering\footnotesize
\setlength{\tabcolsep}{4pt}
\begin{tabular}{llccccc}
\toprule
Setting & Position & 2.5 & Median & 97.5 & Splits & Support \\
\midrule
RAMDocs & support (pooled $\times 2$) & 0.080 & 0.330 & 0.536 & 1{,}000 & 14{,}549 \\
 & mislead1 & $-$0.137 & 0.104 & 0.343 & 1{,}000 & 2{,}561 \\
 & noise1 & $-$0.107 & 0.093 & 0.279 & 1{,}000 & 1{,}888 \\
RAGuard & support (pooled $\times 2$) & 0.504 & 0.658 & 0.766 & 1{,}000 & 3{,}353 \\
 & mislead1 & 0.575 & 0.713 & 0.806 & 1{,}000 & 2{,}240 \\
 & noise1 & 0.187 & 0.377 & 0.530 & 1{,}000 & 1{,}179 \\
RAMDocs forced & support (pooled $\times 2$) & 0.039 & 0.257 & 0.442 & 1{,}000 & 1{,}474 \\
 & mislead1 & $-$0.025 & 0.594 & 0.734 & 950 & 5{,}125 \\
 & noise1 & $-$0.164 & 0.048 & 0.275 & 1{,}000 & 264 \\
\bottomrule
\end{tabular}
\caption{\textbf{Per-position conditional signed stability.} Support:
dual-nonzero cells pooled over splits. Forced mislead1 is undefined on 50
splits (zero support in a half).}
\label{tab:perposition}
\end{table}

\section{Stable-world null calibrations}
\label{app:nulls}

\textbf{Internal arm (sparsity calibration).} The stable world assumes a
fixed per-pair conflict propensity $p_{ij}$ across queries while preserving
the real sparsity mask and each pair's total conflict count. The primary
variant permutes observed conflict indicators within each reader pair
(conditional Monte Carlo test; 5{,}000 simulations, observed 0.138 on 3{,}206
dual-nonzero cells).

\begin{table}[h]
\centering\footnotesize
\begin{tabular}{lccc}
\toprule
Null variant & Null median & $p_{\le \mathrm{observed}}$ & $z$ \\
\midrule
Permutation (primary) & 0.363 & $2\!\times\!10^{-4}$ & $-$15.4 \\
Posterior predictive (beta--binomial EB) & 0.383 & 0.018 & $-$2.30 \\
Plugin EB (variance-shrunk) & 0.261 & 0.131 & $-$1.14 \\
Raw per-pair rates (upper bound) & 0.517 & $2\!\times\!10^{-4}$ & $-$5.20 \\
\bottomrule
\end{tabular}
\caption{\textbf{Stable-world calibrations of the internal signed geometry.}
The EB fit is Beta(21.0, 42.1) over the pooled conflict rate 0.333, with
between-pair variance $\tau^2 = 0.0035$ (stable between-pair SD 0.059).}
\label{tab:loo_nulls}
\end{table}

The calibration also bounds what reliability is \emph{measurable} at our
sample sizes: under the stable null, the expected split-half $\rho$ grows
from 0.104 at 20 queries to 0.191 (40), 0.269 (60), 0.373 (100), 0.463 (140),
and 0.548 (200); 400 simulations per size. The internal arm's signed 0.138 at
100 queries is far below its 0.373 stable-world expectation, while the
ordinal 0.599 exceeds it.

\textbf{PRISM artifact controls} (200 null worlds $\times$ 200 splits each).
Identity null (document identities resampled, chosen-list lengths preserved):
RBO 0.314 [0.021, 0.545]; Jaccard 0.314 [0.050, 0.523]---against the observed
0.786. Order null (chosen sets preserved, order shuffled): RBO 0.703 [0.611,
0.771]; Jaccard 0.742 (invariant by construction). Exact size matching:
median aggregation degenerates (equal-length short lists quantize RBO to a
single constant); the mean-aggregated size-matched geometry is
\emph{more} stable than the raw one (0.862).

\textbf{External arms.} The primary signed calibration preserves the observed
dual-nonzero support and conflict count within every reader-pair$\times$
evidence-position stratum, then permutes the conflict indicators across
queries inside that stratum (2{,}000 simulations). The pair-only variant drops
the position stratification; the ordinal control swaps the two support
positions within reader--query cells (300 simulations).

\begin{table}[h]
\centering\footnotesize
\setlength{\tabcolsep}{4pt}
\resizebox{\linewidth}{!}{%
\begin{tabular}{llcccc}
\toprule
Setting & Null & Null median [2.5, 97.5] & Observed & $p$ & $z$ \\
\midrule
RAMDocs & pair $\times$ position (signed) & 0.376 [0.365, 0.388] & 0.345 & $5\!\times\!10^{-4}$ & $-$5.2 \\
 & pair only (signed) & 0.357 [0.346, 0.368] & 0.345 & 0.024 & $-$2.0 \\
 & type shuffle (ordinal) & 0.763 [0.727, 0.795] & 0.833 & 0.0033 & --- \\
RAGuard & pair $\times$ position (signed) & 0.814 [0.808, 0.820] & 0.748 & $5\!\times\!10^{-4}$ & $-$20.4 \\
 & pair only (signed) & 0.813 [0.806, 0.819] & 0.748 & $5\!\times\!10^{-4}$ & $-$19.6 \\
 & type shuffle (ordinal) & 0.622 [0.577, 0.671] & 0.685 & 0.0033 & --- \\
RAMDocs forced & pair $\times$ position (signed) & 0.516 [0.500, 0.528] & 0.479 & $5\!\times\!10^{-4}$ & $-$4.9 \\
\bottomrule
\end{tabular}
}
\caption{\textbf{External-arm null calibrations.} Signed nulls are one-sided
$p_{\le}$; type-shuffle nulls are $p_{\ge}$ (does instance-level ordering
exceed type-level structure?). The K$=$4 design leaves only the two support
positions swappable, so type-shuffle tests instance-level support ordering
only.}
\label{tab:ext_nulls}
\end{table}

\section{Test--retest and decoding-noise bounds}
\label{app:retest}

\textbf{Internal arm (exact duplicate).} All 1{,}000 conditions of one local
reader (Qwen3.5-9B, deterministic decoding) were accidentally run twice. Of
800 utility cells, 122 are informative (nonzero in either run); the
informative sign-change rate is $f = 0.123$; when both runs are nonzero the
sign holds in 107/109 (0.982). Answer agreement 0.960; utility magnitude
Pearson 0.977. Under independent per-cell flips, observed $\approx$ true
$\times (1-2f)^2 = 0.569 \times$ true, capping true signed reliability at
$0.138/0.569 = 0.242$---below the stable null (0.363) and far below ordinal
(0.599). The run-vs-run distance matrices correlate at 0.786 with mean
$|\Delta d| = 0.020$. Scope: one local reader; API readers ran deterministic
decoding and were not re-run, so a multi-reader repetition remains the gold
standard.

\textbf{External arms (planned triplicate).} 120 stratified queries (60 per
dataset) $\times$ 3 total passes. RAMDocs pools to $f = 0.104$ (546/5{,}257
informative cells; attenuation 0.628; max compatible true signed 0.550): noise
cannot bridge the ordinal--signed gap (0.833 vs.\ 0.345), though it could in
principle account for the small signed-vs-null shortfall (null 0.376).
RAGuard pools to $f = 0.216$ (649/3{,}004; attenuation 0.323), leaving the
bound uninformative (ceiling $>1$). Per-reader $f$ spans 0.015--0.373
(RAMDocs) and 0.000--0.664 (RAGuard), highest for the two reasoning-oriented
endpoints (K3, GPT-5.6-Luna), consistent with their less constrained
decoding. Since decoding noise attenuates observed stability, it works
\emph{against}---never for---the stable signed geometry we report on RAGuard.

\section{Forced-choice label strata}
\label{app:labels}

\begin{table}[h]
\centering\footnotesize
\setlength{\tabcolsep}{4pt}
\resizebox{\linewidth}{!}{%
\begin{tabular}{lcccc}
\toprule
Object & Pooled & gold $=$ A & gold $=$ B & Label-balanced \\
\midrule
Overall signed & 0.479 [0.290, 0.624] & 0.418 [0.123, 0.571] & $-$0.011 [$-$0.132, 0.158] & 0.482 [0.283, 0.629] \\
Support & 0.257 [0.039, 0.442] & 0.156 [$-$0.109, 0.273] & 0.032 [$-$0.147, 0.193] & 0.264 [0.041, 0.472] \\
Mislead1 & 0.594 [$-$0.025, 0.734] & 0.599 [$-$0.021, 0.738] & degenerate & $=$ gold $=$ A \\
Noise1 & 0.048 [$-$0.164, 0.275] & 0.151 [$-$0.229, 0.492] & $-$0.042 [$-$0.290, 0.290] & 0.108 [$-$0.407, 0.652] \\
\bottomrule
\end{tabular}
}
\caption{\textbf{Forced-choice signed stability by label stratum} (median
[2.5, 97.5] over 1{,}000 splits; mislead1 defined on 950). Balanced $=$ mean
of the two stratum distances per split. Noise1 has only 264 dual-nonzero
cells (low power).}
\label{tab:labels}
\end{table}

Two findings. First, the aggregate forced-choice elevation is not a
label-mixing artifact: balancing the strata changes overall signed stability
from 0.479 to 0.482. Second, the mislead1 geometry is measurable only in the
gold\,$=$\,A stratum. When gold\,$=$\,B (the misleading answer sits at A),
readers flip to the misleading letter nearly uniformly; with conflict rates
constant across pairs the stratum contributes no measurable geometry. Uniform
flipping is itself maximal signed agreement, so both strata indicate far more
consistent signed behavior than the open-ended 0.104---the asymmetry marks
\emph{where} cross-reader variance lives, not instability. Closed-book forced
accuracy spans 0.61--0.87 across readers, which limits headroom for
support-side utility.

\section{Transfer experiment details}
\label{app:transfer}

Design: for each of 9 target readers and 50 pre-registered queries (25 NQ, 25
HotpotQA), every source reader's measured utilities induce a tie-aware
preference set (positive-utility documents in retrieval order; rank-based
backfill only to pad), evaluated on the target reader: 187 unique transfer
cells per target, 4{,}050 source $\times$ target $\times$ query cells in
total.

\begin{table}[h]
\centering\footnotesize
\setlength{\tabcolsep}{4pt}
\begin{tabular}{llcc}
\toprule
Target & Nearest source & Nearest regret & Random-source mean regret \\
\midrule
DeepSeek-V4-Flash & Qwen3.6-Flash & 0.050 & 0.054 \\
GLM-5.2 & Qwen3.6-Flash & 0.024 & 0.027 \\
GPT-5.6-Luna & Qwen3.6-Flash & 0.017 & 0.031 \\
K3 & Qwen3.6-Flash & $-$0.013 & 0.003 \\
Llama-3.1-8B & Qwen3.6-Flash & 0.026 & 0.044 \\
Llama-3.3-8B & Qwen3.6-Flash & 0.021 & 0.025 \\
Ministral-8B & Llama-3.1-8B & $-$0.006 & 0.009 \\
Qwen3.5-9B & Qwen3.6-Flash & 0.027 & 0.036 \\
Qwen3.6-Flash & DeepSeek-V4-Flash & 0.023 & 0.051 \\
\bottomrule
\end{tabular}
\caption{\textbf{Nearest-source vs.\ random-source transfer regret} (F1 with
the target's own selection minus F1 with the transferred selection, averaged
over 50 queries). Nearest source by behavior-profile distance under the
saturated probe bank (Appendix~\ref{app:probes}). Negative regret means the
transferred set beat the target's own set---possible because the target's own
selection is itself sparse and backfilled.}
\label{tab:transfer}
\end{table}

Noise floor: 47.7\% of source pairs select identical document sets for a
query (mean Jaccard 0.421 among the rest; 3.74 distinct sets per query);
mean $|$regret$|$ is 0.035 F1 against a mean per-cell standard error of
0.037; 54.4\% of cells have zero cross-source spread; selections average 1.08
positive-utility documents against 2.77 backfilled. Predictability:
behavior-profile distance vs.\ symmetrized regret $\rho = 0.047$
(reader-label permutation $p = 0.901$, 10{,}000 permutations); in-sample
oracle utility distance vs.\ regret $\rho = -0.271$ ($p = 0.264$); per-target
directional tests average $\rho = 0.335$ but pool to $-0.050$ (56 cells).
Split-half reliability of the regret matrix itself: $-0.281$ [$-0.530$,
0.001] over 200 splits. The frozen decision rule ($\rho \ge 0.30$,
$p < 0.05$, every nearest source $\ge 10\%$ regret reduction, nearest
direction no worse on both datasets) was not met; per-dataset nearest $-$
random regret is $-0.017$ (HotpotQA) and $-0.008$ (NQ).

\section{Behavioral probe bank (uninformative)}
\label{app:probes}

An earlier stage of this project probed readers with a 48-pair fictional
behavior bank (six dimensions) intended to predict utility geometry from
behavioral profiles. On the finalized 9-reader roster the bank saturates: 7 of
9 readers answer all pairs at ceiling (the two newest API readers at EM
$=1.0$), leaving nonzero behavioral variation for only two local readers. The
behavior--geometry association on this roster is $\rho = 0.265$ (permutation
$p = 0.424$)---not estimable with a saturated instrument. We report this as a
measurement-ceiling observation, not as evidence about the behavior--utility
relationship: resolving modern readers requires a harder probe bank, which we
did not build. No main-text claim depends on the probe bank.

\section{Artifact map}
\label{sec:artifacts}

\begin{table}[h]
\centering\footnotesize
\setlength{\tabcolsep}{4pt}
\resizebox{\linewidth}{!}{%
\begin{tabular}{p{5.4cm}p{8.6cm}}
\toprule
Claim & Artifact (filename under \texttt{artifacts/\allowbreak{}}) \\
\midrule
Controlled heterogeneity (33.3\% sign conflict; 72.3\% activity asymmetry) &
\texttt{reader\_\allowbreak{}geometry\_\allowbreak{}stability/\allowbreak{} heterogeneity\_\allowbreak{}summary\_\allowbreak{}v3\_\allowbreak{}9readers.json} \\
Variance decomposition (29.8\% reader$\times$query; 68\% reader-involving sum) &
\texttt{reader\_\allowbreak{}geometry\_\allowbreak{}stability/\allowbreak{} variance\_\allowbreak{}decomposition\_\allowbreak{}v3\_\allowbreak{}9readers.json} \\
LOO ordinal 0.599 / signed 0.138 &
\texttt{reader\_\allowbreak{}geometry\_\allowbreak{}stability/\allowbreak{} reader\_\allowbreak{}geometry\_\allowbreak{}stability\_\allowbreak{}v3\_\allowbreak{}9readers.json} \\
Sparsity calibration (null 0.363) &
\texttt{reader\_\allowbreak{}geometry\_\allowbreak{}stability/\allowbreak{} sparse\_\allowbreak{}reliability\_\allowbreak{}calibration\_\allowbreak{}v3\_\allowbreak{}9readers.json} \\
Matched metrics (0.650 vs.\ 0.138) &
\texttt{reader\_\allowbreak{}geometry\_\allowbreak{}stability/\allowbreak{} matched\_\allowbreak{}metric\_\allowbreak{}robustness\_\allowbreak{}v3\_\allowbreak{}9readers.json} \\
Test--retest bound ($f = 0.123$) &
\texttt{reader\_\allowbreak{}geometry\_\allowbreak{}stability/\allowbreak{} test\_\allowbreak{}retest\_\allowbreak{}qwen3\_\allowbreak{}5\_\allowbreak{}v3\_\allowbreak{}9readers.json} \\
PRISM ordinal 0.786 + nulls &
\texttt{prism\_\allowbreak{}stability/\allowbreak{}prism\_\allowbreak{}ordinal\_\allowbreak{}stability\_\allowbreak{}v1.json}; \texttt{prism\_\allowbreak{}ordinal\_\allowbreak{}stability\_\allowbreak{}nulls\_\allowbreak{}v1.json} \\
RAMDocs/RAGuard geometry, nulls, per-position &
\texttt{signed\_\allowbreak{}replication/\allowbreak{}analysis/\allowbreak{} signed\_\allowbreak{}replication\_\allowbreak{}geometry\_\allowbreak{}v2\_\allowbreak{}13readers.json} \\
Forced-choice $\Delta$ + per-position + null &
\texttt{signed\_\allowbreak{}replication\_\allowbreak{}forced/\allowbreak{}analysis/\allowbreak{} forced\_\allowbreak{}choice\_\allowbreak{}v2\_\allowbreak{}13readers.json} \\
Label-balanced sensitivity &
\texttt{signed\_\allowbreak{}replication\_\allowbreak{}forced/\allowbreak{}analysis/\allowbreak{} label\_\allowbreak{}balanced\_\allowbreak{}sensitivity\_\allowbreak{}v2\_\allowbreak{}13readers.json} \\
Transfer experiment (4{,}050 cells) &
\texttt{gate15b/\allowbreak{}gate15b\_\allowbreak{}summary\_\allowbreak{}v3\_\allowbreak{}9readers.json} \\
Oracle/regret post-mortem &
\texttt{gate15b/\allowbreak{}gate15b\_\allowbreak{}postmortem\_\allowbreak{}v3\_\allowbreak{}9readers.json} \\
\bottomrule
\end{tabular}
}
\caption{Primary claims and their frozen artifacts.}
\end{table}

\end{document}